%% file: main.tex
\documentclass[conference]{IEEEtran}
\IEEEoverridecommandlockouts

\usepackage{cite}
\usepackage{hyperref}
\usepackage{amsmath}
\usepackage{amssymb}
\usepackage{amsthm}
\usepackage{booktabs}
\usepackage{subcaption}
\usepackage{graphicx}
\usepackage{float}

\newtheorem{innercustomgeneric}{\customgenericname}
\providecommand{\customgenericname}{}
\newcommand{\newcustomtheorem}[2]{%
  \newenvironment{#1}[1]
  {%
   \renewcommand\customgenericname{#2}%
   \renewcommand\theinnercustomgeneric{##1}%
   \innercustomgeneric
  }
  {\endinnercustomgeneric}
}

\newcustomtheorem{customthm}{Theorem}
\newcustomtheorem{customlemma}{Lemma}

\newcommand{\vecX}{\mathbf{x}}

\newcommand{\vecZ}{\mathbf{z}}

\newcommand{\vecM}{\boldsymbol{\mu}}

\newcommand{\id}{i}

\DeclareMathOperator*{\argmax}{argmax}

\input{preamble}

\definecolor{cvprblue}{rgb}{0.21,0.49,0.74}

\title{Holistic Uncertainty Estimation For Open-Set Recognition}

\author{Erlygin Leonid\\
Skoltech, Moscow, Russia\\
{\tt\small L.Erlygin@skoltech.ru}
\and
Alexey Zaytsev\\
Skoltech, Moscow, Russia\\
{\tt\small A.Zaytsev@skoltech.ru}
}

\begin{document}
\maketitle
\input{parts/abstract}
\input{parts/introduction}
\input{parts/related_work}
\input{parts/methodology/problem_statement}
\input{parts/methodology/compared_methods}
\input{parts/methodology/mc_theory}

\input{parts/experiments}

\input{parts/conclusion}


\bibliographystyle{ieeetr}
\bibliography{main}

\input{parts/appendix}
\end{document}

%% file: preamble.tex
\usepackage[dvipsnames]{xcolor}


%% file: parts/abstract.tex
\begin{abstract}
Accurate uncertainty estimation is a critical challenge in open-set recognition, where a probe biometric sample may belong to an unknown identity. 
It can be addressed through sample quality estimation via probabilistic embeddings.
However, the low variance of probabilistic embedding only partly implies a low identification error probability: an embedding of a sample could be close to several classes in a gallery, thus yielding high uncertainty despite high sample quality.

We propose \textit{HolUE} --- a holistic uncertainty estimation method based on a Bayesian probabilistic model; it is aware of two sources of ambiguity in the open-set recognition system: (1) the gallery uncertainty caused by overlapping classes and (2) the uncertainty of embeddings.
Challenging open-set recognition datasets, such as IJB-C for the image domain and VoxBlink for the audio domain, serve as a testbed for our method. 
We also provide a new open-set recognition protocol for the identification of whales and dolphins.  
In all cases, HolUE better identifies recognition errors than alternative uncertainty estimation methods, including those based solely on sample quality.
\end{abstract}

%% file: parts/introduction.tex
\section{Introduction}
\label{sec:introduction}


The open-set recognition (OSR) problem arises in many practical applications~\cite{Gnther2017TowardOF}.
The key distinction between the open-set and closed-set recognition problem is the possibility of occurrence of previously unseen subjects.
During the inference, the OSR system must accept known subjects, identify them by finding the most similar subjects in the gallery of known people, and reject subjects from outside the gallery, i.e., those not enrolled in the gallery.
An established baseline solution to OSR is to train an embedding model that produces discriminative feature vectors and uses cosine distance between them to estimate the semantic difference between samples~\cite{deng2019arcface, face_handbook}.
The probe sample is rejected if its distance to each subject in the gallery is higher than a certain threshold. 
Otherwise, it is accepted, and the closest gallery subject label is assigned.
This approach strongly relies on the discriminative power of the embedding model. 
However deterministic embeddings are not robust enough to deal with corrupted probe samples.
In case of the image domain, the embeddings can shift unpredictably when some visual features are missing, leading to recognition errors~\cite{pfe}.
To address this, probabilistic embedding models, such as $\mathrm{PFE}$~\cite{pfe} and $\mathrm{SCF}$~\cite{scf}, have gained attention. 
Probabilistic embeddings enable uncertainty-aware distance computation, improve template aggregation by producing a single embedding from multiple samples of the same person and facilitating low-quality sample filtering~\cite{pfe, Kail2022ScaleFaceUD}.

\begin{figure} 
    \includegraphics[width=0.45\linewidth]{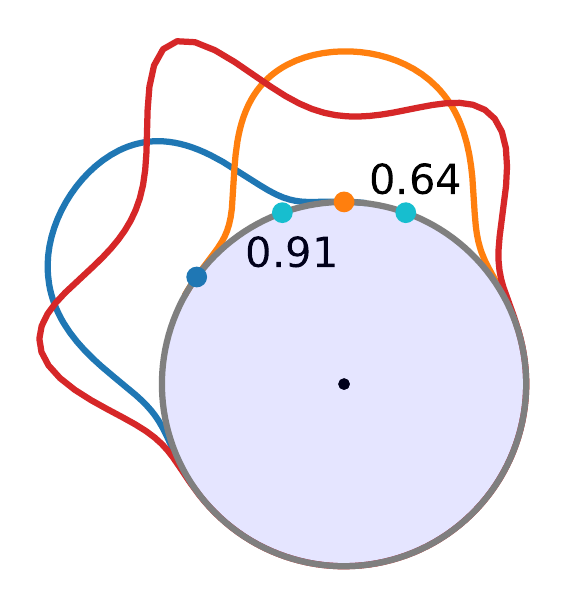}
    \centering
    \caption{A toy example in two-dimensional space showcasing the importance of the information about embeddings relative position.
    Embeddings are normalized to have a unit norm.
    The gallery consists of orange and dark blue classes, and the corresponding curves show the posterior class probability given by our method $\mathrm{GalUE}$.
    The red curve represents HolUE combined uncertainty.
    Cyan points represent the embedding of probe sample: both points have the same distance to the closest orange class, but the left point has a higher uncertainty $0.91$ c.t. $0.64$ because it is also close to the dark blue class.}
    \label{fig:false_ident_toy}
\end{figure}

The existing OSR literature typically overlooks uncertainty prediction, focusing primarily on reducing recognition errors. 
We suspect this happens because existing uncertainty estimates lack the essential properties needed for a robust OSR uncertainty estimator.
Uncertainty estimation using probabilistic embeddings to measure sample quality is reasonable, but it does not cover all causes of OSR errors.
For instance, an OSR system should exhibit high uncertainty when embedding of a probe sample close to several subjects in the gallery. 
In such cases, the probabilistic embedding uncertainty predicted by $\mathrm{SCF}$~\cite{scf} can be relatively low and fails to reflect the uncertainty related to identification ambiguity adequately.
Joining probe sample quality estimate and information about embeddings' relative position to build a unified uncertainty estimate necessitates dealing with complex Bayesian integrals, and this issue hasn't been resolved yet.

Our work introduces the holistic approach to estimating uncertainty for OSR.
Based on possible error types, the method assigns low confidence in three cases: low sample quality, samples' embedding is close to several gallery classes, 
and embedding is near the decision boundary between rejection and acceptance.
An illustration of the second case is presented in Figure~\ref{fig:false_ident_toy}.
To consider information about embeddings' relative position, we reformulate the open-set recognition problem as a classification problem by adding an out-of-gallery class corresponding to unknown subjects.
We define the Bayesian model to get the categorical distribution of possible decisions in the OSR system for each probe sample.
In such a formulation, the class with the maximum probability is predicted, and the probability of this class is used as a measure of recognition uncertainty.
Then, to incorporate information about sample quality, we consider the embedding distribution predicted by $\mathrm{SCF}$~\cite{scf}. 
The most complete declaration of presented uncertainties is the posterior class probability represented as the integral over the embedding space.
We calculate KL-divergence between this posterior and uniform prior class distribution to get an uncertainty estimate of the probe.
Uncertainty is high when the posterior is close to the uniform distribution.

To our knowledge, this work is the first to combine information about embeddings' relative position with information about embeddings' quality to obtain a robust uncertainty estimation method in OSR.

The key contributions of our work are as follows:
\begin{itemize}
\item A gallery-aware probabilistic method $\mathrm{GalUE}$ for uncertainty estimation in an open-set recognition problem. 
It produces decision probabilities that incorporate information about embeddings relative positions.
\item A holistic method $\mathrm{HolUE}$ to account for all cases of prediction ambiguity.
In this method, we combine the $\mathrm{GalUE}$ approach with a neural network-based sample quality estimator $\mathrm{SCF}$ via principled Bayesian model of OSR system.
\item A new Open Set Recognition dataset \textit{Whale} based on Happywhale dataset~\cite{happy-whale-and-dolphin} to test if our method can be used in a broader than face domain~range.
\item A careful experimental evaluation of the proposed approach. 
Using four datasets, IJB-B~\cite{ijbb}, IJB-C~\cite{ijbc},  VoxBlink~\cite{voxblink}, and Whale, we show that our method outperforms existing uncertainty estimators based on a partial understanding of the uncertainty of a recognition model. 
\end{itemize}

%% file: parts/related_work.tex
\section{Related Work}
\label{sec:related_work}
\paragraph{Open-set recognition problem}
The amount of conducted research in open-set recognition (OSR) demonstrates the importance of this topic and specific properties of the problem~\cite{1811.08581}.
The OSR task differs from the classification task because previously unseen (out-of-gallery) classes are present during inference.
Existing works have focused on developing OSR algorithms that can accurately reject out-of-gallery probe samples, leaving the uncertainty estimation out of scope. 
Current solutions to the OSR problem for the face and audio domains are no exception: most works use simple cosine distance-based rejection rule and focus instead on the improvement of the discriminative power of the embedding model by incorporating recent advances in the field, including deep learning models trained with utilization of loss functions like ArcFace~\cite{deng2019arcface} and CosFace~\cite{Wang_2018_CVPR}.


\paragraph{Uncertainty estimation in open-set recognition}
Uncertainty estimation can provide benefits for recognition problems --- both in terms of the final recognition accuracy and 
decision-making based on the estimated level of confidence~\cite{pfe}.

Uncertainty is typically divided by source to model and data ones~\cite{ABDAR2021243}.
Model uncertainty is caused by imperfect embedding model training and limited training data samples. 
This may result in noticeable embedding vector variation for different data samples of the same person, leading to OSR errors.
In our study, we use a pre-trained embedding model, leaving the task of pure model uncertainty estimation in OSP for further research. 
Data uncertainty describes uncertainty in input samples, which can be caused by sample corruption~\cite{pfe}.
There are three types of OSR errors: false acceptance, false rejection, and misidentification~\cite{face_handbook}, and an ideal recognition system should predict high uncertainties when it is likely to make any of these errors.
Although uncertainty in OSR has been studied~\cite{transductive_conf, osr_2012}, to our knowledge, it was never used as the likelihood of OSR errors.
Instead, its usage focused on finding out-of-gallery samples.
\paragraph{Probabilistic embeddings}
Numerous works are dedicated to predicting probabilistic sample embeddings~\cite{pfe, scf, Kail2022ScaleFaceUD}.
In contrast to deterministic recognition approaches, such as ArcFace~\cite{deng2019arcface}, they predict the whole distribution of embeddings.
The variance of the predicted distribution can be used to measure nose level in samples' features.
Samples of low quality would have high variance and high data uncertainty.
These ambiguous samples can lead to OSR errors. 
Thus, their identification can help to make recognition more robust.
In the classification task, data uncertainty is often described as the entropy of class distribution, conditional on features. 
If we formulate the OSR problem as a classification task, then the data uncertainty of the OSR system can be defined similarly.
Uncertainty estimation methods that use only sample quality scores do not consider the intrinsic structure of classes in the OSR system, and thus, their uncertainty scores are incomplete.  

\paragraph{Research gap}
To our knowledge, no present method has used the data uncertainty of an OSR probabilistic model together with the uncertainty caused by samples.
We construct a Bayesian model of embedding distribution of known subjects, which yields OSR decision probabilities and natural data uncertainty estimates.
Combining this uncertainty with data uncertainty produced by the probabilistic embedding model, we get an uncertainty estimator that can cover all sources of possible OSR errors.

%% file: parts/methodology/problem_statement.tex
\section{Methods} 
\label{sec:methodology}
This section is divided into two parts: an introduction and a description of our proposed method.
In \emph{the introductory part}, we define the open-set recognition (OSR) problem, following notation from~\cite{ISO/IEC2382-37} biometrics standard.
We also discuss the task of uncertainty estimation in OSR and provide an example baseline solution to OSR.
\emph{The second part} describes our proposed method, which provides holistic uncertainty estimation for OSR, detailing why it covers the gaps in existing techniques.

\subsection{Open-set recognition problem statement}
\label{sec:problem_statement}
An OSR system has access to a gallery $\mathcal{G}$ of biometric samples of subjects.
The gallery is structured as a set of \textit{templates} $\mathcal{G} = \{g_1, \dots, g_K\}$ of size $K$.
Each template $g_i$ consists of available biometric samples of a subject with index $i$.
So, the gallery has $K$ distinct classes of samples associated with known subjects. 
Further on, we will use the words "class" and "subject" interchangeably.  

At inference time, the system receives a probe $p$ with several biometric samples of a person $\{\vecX_1, \ldots, \vecX_m \}$ with, in most cases, $m = 1$, and performs the biometric search.
The system should correctly answer two questions:
\begin{enumerate}
    \item Does a template mated with the probe $p$ exist in the gallery $\mathcal{G}$?
    \item If such a template $g^*$ exists, then what is its identifier $\hat{\id}(g^*) \in \{1, \ldots, K\}$?
\end{enumerate}
In other words, the system decides whether the person represented as $p$ is known.
If the decision is positive, $p$ is \textit{accepted}, and a class is assigned. 
Otherwise, $p$ is \textit{rejected} as a non-mated probe.

\subsubsection{Uncertainty in open-set recognition system}
A robust OSR system should be capable of estimating the uncertainty of its predictions. 
High uncertainty should correlate with a high probability of error, prompting the system to \textit{filter}, i.e. refuse to recognize the probe $p$ to avoid potential false acceptance or rejection.
In applications, the biometric system may suggest that the user retake the biometric sample and try again.

In OSR, we identify two sources of uncertainty.
Firstly, uncertainty arises when the biometric features in the presented sample are noisy or corrupted.
Additionally, the system experiences uncertainty in its recognition decision when the embedding of the probe is close to multiple classes in the gallery or lies on the decision boundary between acceptance and rejection. 
For instance, when two identities with similar biometric features are present, it becomes unclear which class should be assigned to $p$ if it is accepted.

Uncertainty estimation can be formalized by introducing \emph{quality score} $q(p)$ that takes a probe $p$ as an input. 
If $q(p)$ is lower than a threshold $t$, the system asks the user to provide new biometric identifier, hoping to decrease the overall recognition error rate.  

\subsubsection{Baseline OSR method}
\label{sec:pairwise_distance}
The simplest OSR method involves the computation of \textit{acceptance score} $s(p)$, where $s(*)$ is a decision function.
If $s(p)$ exceeds a threshold $\tau$, the probe $p$ is accepted and classified.
A common decision function is the cosine similarity with the closest class in gallery~\cite{face_handbook}:
\begin{equation*}
    s(p) = \max_{c \in \{1, \dots, K\}} \vecM^T_{c} \vecZ,
\end{equation*}
where $\vecZ = \vecZ(p)$ is the aggregated embedding of $p$;
$\vecM_c$ is the aggregated embedding of a template $g_c$.
Low values of $s(p)$ let the system assume that the probe is non-mated with every known subject and thus has to be rejected.
In this way, we obtain a two-step algorithm:
\begin{enumerate}
    \item If $s(p) \geq \tau$, $p$ is accepted.
    \item If $p$ is accepted, $\hat{\id}(p) = \argmax_{c \in \{1, \dots, K\}} \vecM^T_{c} \vecZ$.
\end{enumerate}

For this OSR method, we propose an ad-hoc quality measure, which utilizes acceptance score $s(p)$:
\begin{equation*}
    q_{\mathrm{AccScr}}(p) =  |s(p) - \tau|.
\end{equation*}
A similar measure was used to estimate uncertainty in the face verification system~\cite{Huber_2022_BMVC}, where a binary classification task is considered. 
The intuition behind this approach comes from the fact that the recognition system will likely make false acceptance or rejection errors for probe $p$ with an acceptance score near the decision threshold~$\tau$.
The idea of $q_{\mathrm{AccScr}}$ score is close to the approach of using the maximum class probability as the confidence score~\cite{kotelevskii2022nonparametric}.

%% file: parts/methodology/compared_methods.tex
\subsection{Holistic uncertainty estimation}
\label{sec:methods}
In this section, we define our method, HolUE (Holistic Uncertainty Estimate) -- a Bayesian probabilistic model of the OSR system, which gives an uncertainty estimate aware of two sources of ambiguity. 
We consider the problem of the reconstruction of the class distribution $p(c|\vecX)$ given a single-sample probe $p = \{\vecX\}$.
This distribution can be computed with the following integral:
\begin{equation}\label{eq:z_marginalization}
    p(c | \vecX) = \int_{\mathbb{S}_{d - 1}} p(c|\vecZ) p(\vecZ | \vecX) d \vecZ,
\end{equation}
where $c$ is a class label, and $\vecZ \in \mathbb{S}_{d - 1}$ is the embedding of a biometric sample $\vecX$ that lies on a $d$-dimensional unit sphere $\mathbb{S}_{d - 1}$.
So, $p(\vecZ | \vecX)$ is the distribution of the embeddings given $\vecX$, and $p(c|\vecZ)$ is the distribution of the class given the embedding.
The entropy of $p(c | \vecX)$ would correspond to the uncertainty of the model.

Both terms in the integral reflect different facets of uncertainty: $p(\vecZ | \vecX)$ provides information about uncertainty related to sample quality; $p(c|\vecZ)$ gives a measure of uncertainty related to gallery ambiguity.
To complete the method, we should define approaches to estimate both these conditionals and to calculate the integral.
Below, we describe a way to model $p(c | \vecZ)$ to obtain gallery-aware method $\mathrm{GalUE}$. 
Then, we propose an approach to get distribution $p(\vecZ | \vecX)$, which incorporates information about biometric sample quality.
Finally, we approximately compute $p(c|\vecX)$ using \eqref{eq:z_marginalization} to obtain method $\mathrm{HolUE}$ that takes into account both sources of uncertainty.


\subsubsection{GalUE: Bayesian model for deterministic embeddings}
\label{sec:deterministic_embedding}
When the embedding $\vecZ$ of sample $\vecX$ is deterministic, e.g., predicted by an Arcface-trained encoder~\cite{deng2019arcface}, the Bayesian model of the OSR system can be constructed.
We model $p(c|\vecZ)$ and use it to make predictions and compute uncertainty.
In our work, generative approach is taken, so $p(\vecZ | c)$ and $p(c)$ are defined to calculate $p(c | \vecZ)$, with the help of Bayes' rule:
\begin{equation}
\label{eq:bayes_formula}
    p(c | \vecZ) = \frac{p(\vecZ | c) p(c)}{p(\vecZ)}.
\end{equation}

\paragraph{Model for $p(c)$}
We assume the class label $c$ is a mixed random variable $c \in \{1, \dots K\}\cup (K, K + 1]$. 
Discrete values of the random variable $c$ correspond to gallery classes, and continuous values from $(K, K + 1]$ represent the continuum of out-of-gallery classes.
Natural assumption for $p(c)$ is the uniform distribution on $c$ with the probability density function:
\begin{equation*}
    p(c) = \frac{1 - \beta}{K} \sum_{i=1}^K \delta(c - i) + \beta \mathrm{I}\{c \in (K, K + 1] \},
\end{equation*}
where $K$ -- number of gallery classes; $\delta$ -- Dirac delta function; $\beta$ -- prior probability of out-of-gallery class; $\mathrm{I}$ -- the indicator function.
We use $\beta=0.5$ in our experiments.
Naturally, in-gallery classes have an equal prior probability.

The necessity of $c$ to be continuous instead of a single number for out-of-gallery classes follows from a closer look into the integral from~\eqref{eq:z_marginalization}.
Let us consider a low-quality in-gallery sample with $\vecX$ whose embedding distribution mean lies far from all gallery classes.
Naturally, its uncertainty should be high to avoid an ungrounded false rejection.
However, if $c$ is discrete, then for such $\vecX$, it holds that $p(\{\text{probe sample is out-of-gallery}\}|\vecX) \approx 1$ and the uncertainty is low.
In this case, the variance of embeddings distribution $p(\vecZ | \vecX)$ will be dismissed, and strong false rejected samples will have low uncertainty. 

\paragraph{Model for $p(\vecZ|c)$}
One can utilize von Mises-Fisher (vMF)~\cite{vmf} distributions defined on a $d$-dimensional unit sphere $\mathbb{S}_{d - 1} \subset \mathbb{R}^d$, to model class conditional embedding distribution $p(\vecZ | c)$ for each gallery class~$c \in \{1, \ldots, K\}$:
\begin{align*}
        p_{\mathrm{vMF}}(\vecZ|c) &= \mathcal{C}_d(\kappa)\exp\left(\kappa\vecM^T_{c} \vecZ \right), \\
        \mathcal{C}_d(\kappa) &= \frac{(\kappa)^{d/2 - 1}}{(2\pi)^{d/2}\mathcal{I}_{d/2-1}(\kappa)},
\end{align*}
where $\vecM_c$ is the aggregated embedding of a gallery template $g_c$; $\kappa$ is a concentration hyperparameter constant for all classes $c$, $\mathcal{I}_{\alpha}$ is a modified Bessel function of the first kind of the order $\alpha$.
In our model, embeddings $\vecM_c^{\circ}$ of out-of-gallery classes are uniformly distributed on $\mathbb{S}_{d - 1}$ and each class $c$ has the corresponding embedding:
\begin{equation}\label{eq:oog_z_distr}
    p(\vecZ|c) = \delta \left(\vecZ - \vecM_c^{\circ}\right),\quad c\in (K,K+1].
\end{equation}

\begin{customlemma}{1}\label{lem:p_z}
With $p(c)$ and $p(\vecZ|c)$ defined above,
    \begin{equation*}
        p(\vecZ) = \frac{1-\beta}{K}\sum_{i=1}^{K}p(\vecZ|c=i) + \frac{\beta}{S_{d-1}},
    \end{equation*}
\end{customlemma}
where $S_{d-1}$ is area of d-dimensional unit sphere.
The proof is in Appendix \ref{proof:p_z}.

Let $p_0$ be the probability of out-of-gallery class and $p_c$ gallery classes probabilities:
\begin{align*}
    p_0 & \equiv \int_{K}^{K + 1} p(c|\vecZ) dc, \\
    p_c &\equiv \mathbb{P}(c|\vecZ), c \in \{1, \dots, K\},
\end{align*}
obviously $p_0 = 1 - \sum_{c = 1}^K p_c$.
The exact formula for the posterior class probabilities $\mathbb{P}(c|\vecZ)$ can be found in Appendix~\ref{sec:computation}.

With this, the probe is rejected if the probability of being an out-of-gallery class is the highest:
\begin{enumerate}
    \item If $p_0>\max_{c\in\{1, \dots, K\}}p_c$, $\vecX$ is rejected. 
    \item If $\vecX$ is accepted, $\hat{\id}(\vecX) = \argmax_{c\in\{1,\dots,K\}}p_c$.
\end{enumerate}
This decision rule is equivalent to the baseline method mentioned in Section \ref{sec:pairwise_distance} with a specific function $f$: 
\begin{equation*}
    \tau = f(\kappa, \beta, K).
\end{equation*}
In the Appendix \ref{sec:score_function_connection}, the exact form of the function $f$ for vMF distribution can be found.
Similarly to the baseline method, where certain $\tau$ gives a particular false acceptance rate, $\kappa$ can be selected such that the recognition system yields the same predictions.

Although predictions of this method are equivalent to the baseline solution with appropriate $\tau$, the proposed approach provides a natural way to estimate the uncertainty of the OSR system.
With access to predicted probabilities for all classes, we can compute the maximum probability uncertainty estimate:
\begin{equation*}\label{eq:unc_vmf}
    q_{\mathrm{GalUE}}(\vecZ) = \max_{c \in \{0, \dots, K\}} p_c.
\end{equation*}
An alternative option, which computes the entropy of this categorical distribution, provided no improvements in our experiments.
We name the proposed gallery-aware uncertainty estimation approach \textit{GalUE}.


\subsubsection{Estimation of embedding distribution}\label{sec:embedding_uncertainty} 

Above, we have considered the case with a common hidden assumption that the biometric sample has perfect quality.
In this case, the embedding model gives us the true feature vector associated with a particular person, and the embedding $\vecZ$ is deterministic.
In practice, samples may have poor quality~\cite{karpukhin2023probabilistic}.
Prominent face recognition methods, $\mathrm{PFE}$~\cite{pfe}, $\mathrm{SCF}$~\cite{scf} and ScaleFace (SF)~\cite{Kail2022ScaleFaceUD} suggest using probabilistic embeddings to capture uncertainty caused by poor sample quality.
In our experiments, we utilize $\mathrm{SCF}$~\cite{scf} model for image and audio domains, and thus: 
\begin{equation*}
p(\vecZ|\vecX) = p_{\mathrm{vMF}}(\vecZ; \boldsymbol{\mu}(\vecX), \kappa(\vecX)),
\end{equation*}
where $p_{\mathrm{vMF}}(\vecZ;\boldsymbol{\mu}(\vecX),\kappa(\vecX))$ -- von Mises-Fisher distribution on $\mathbb{S}_{d - 1}$, and $\boldsymbol{\mu}(\vecX),\kappa(\vecX)$ -- mean and concentration for the sample $\vecX$.
The concentration value $\kappa(\vecX)$ of $\mathrm{SCF}$, variance vector $\boldsymbol{\sigma}^2(\vecX)$ predicted by $\mathrm{PFE}$ and scale $s(\vecX)$ in ArcFace loss predicted by $\mathrm{SF}$ also can be directly used as quality measure:
\begin{align*}
    &q_{\mathrm{SCF}}(\vecX) = \kappa(\vecX),\\
    &q_{\mathrm{PFE}}(\vecX) = -H(\boldsymbol{\sigma}^2(\vecX))\\
    &q_{\mathrm{SF}}(\vecX) = s(\vecX)
\end{align*}
where $H$ - harmonic mean.
We use this uncertainty estimation methods based on quality of a sample as another baseline.

%% file: parts/methodology/mc_theory.tex
\subsubsection{HolUE: combined uncertainty estimate}\label{sec:combined_ue}
Now we can use $p(c|\vecZ)$ and $p(\vecZ|\vecX)$ defined above to compute the integral~\eqref{eq:z_marginalization} and get a combined uncertainty estimate.
Using ~\eqref{eq:bayes_formula} and (\ref{eq:oog_z_distr}), we get $p(c|\vecX)$ for out-of-gallery classes, $c\in (K, K + 1]$:
\begin{equation*}
    p(c|\vecX) = \int_{\mathbb{S}_{d-1}}\frac{\delta(\vecZ-\vecM_c^{\circ})\beta}{p(\vecZ)}p(\vecZ|\vecX)d\vecZ = \beta\frac{p(\vecM_c^{\circ}|\vecX)}{p(\vecM_c^{\circ})}.
\end{equation*}
where $\vecM_c^{\circ}$ is the embedding of class $c$.

For in-gallery classes with $c\in\{1, \dots, K\}$:
\begin{equation}\label{eq:gallery_prob}
    \mathbb{P}(c|\vecX) = \int_{\mathbb{S}_{d - 1}} \frac{1 - \beta}{K}\frac{p(\vecZ|c)}{p(\vecZ)}p(\vecZ|\vecX) d\vecZ.
\end{equation}
Now that we have posterior class distribution, we propose to compute KL-divergence, also known as relative entropy, between $p(c|\vecX)$ and $p(c)$ to measure how much class uncertainty is reduced when a sample $\vecX$ is presented:
\begin{align*}
    &\operatorname{D}_{\mathrm{KL}}(p(c|\vecX) \| p(c)) = \sum_{c = 1}^K \mathbb{P}(c|\vecX) \log \frac{\mathbb{P}(c|\vecX)}{\mathbb{P}(c)} + \\
&+\int_K^{K+1} p(c|\vecX) \log \frac{p(c|\vecX)}{p(c)} dc,\\
& \int_K^{K+1} p(c|\vecX) \log \frac{p(c|\vecX)}{p(c)} d c =\\
& = \int_{\mathbb{S}_{d - 1}} \frac{1}{S_{d - 1}} \beta\frac{p\left(\vecM_c^\circ| \vecX\right)}{p\left(\vecM_c^\circ\right)} \log \frac{\beta p\left(\vecM_c^\circ|\vecX\right)}{\beta p\left(\vecM_c^\circ\right)} d \vecM_c^\circ. 
\end{align*}
In our experiments we approximate $\mathbb{P}(c|\vecX)$ using mean value of $p(\vecZ|\vecX)$, $\vecM_{\vecX}=\vecM(\vecX)$ (See \eqref{eq:gallery_prob}):
\begin{equation*}
    \mathbb{P}(c|\vecX)\approx \frac{1-\beta}{K}\frac{p(\vecM_{\vecX}|c)}{p(\vecM_{\vecX})}.
\end{equation*}
In the same way, we approximate the second term in KL-divergence:
\begin{align*}
\operatorname{D}_{\mathrm{KL}}(p(c|\vecX) \| p(c)) \approx &\sum_{c=1}^K \mathbb{P}(c|\vecX) \log \frac{\mathbb{P}(c|\vecX)}{\mathbb{P}(c)}+ \\
+&\frac{\beta}{S_{d-1}} \frac{1}{p(\vecM_{\vecX})} \log\frac{p\left(\vecM_{\vecX}|\vecX\right)}{p\left(\vecM_{\vecX}\right)},
\end{align*}

\paragraph{Post-processing}
To secure the numerical stability of our approach, we use temperature scaling of posterior class probabilities.
This scaling does not change predictions of the OSR system and retains the order of quality scores. 
\begin{customlemma}{2}\label{lemma:t_scale}
    KL-divergence after scaling with temperature T has the following form:
\begin{align*}
        &\operatorname{D}_{\mathrm{KL}}(p_T(c|\vecX) \| p(c))=\underbrace{\sum_{c=1}^K \mathbb{P}_T(c|\vecX) \log \frac{\mathbb{P}_T(c|\vecX)}{\mathbb{P}(c)}}_{\mathrm{KL}_1}+ \\
+&\underbrace{\frac{\beta^{\frac{1}{T}}}{S_{d-1}^{\frac{1}{T}}} \frac{1}{p(\vecM_{\vecX})} \log\left[\left(\frac{\beta}{S_{d-1}}\right)^{\frac{1}{T}-1}\frac{p\left(\vecM_{\vecX}|\vecX\right)}{p\left(\vecM_{\vecX}\right)}\right]}_{\mathrm{KL}_2}, \\
&\mathbb{P}_T(c|\vecX) = \frac{\left(\frac{1-\beta}{K}\right)^{\frac{1}{T}}p(\vecM_{\vecX}|c)^{\frac{1}{T}}}{\left(\frac{1-\beta}{K}\right)^{\frac{1}{T}}\sum\limits_{c=1}^{K}p(\vecM_{\vecX}|c)^{\frac{1}{T}} + \left(\frac{\beta}{S_{d-1}}\right)^{\frac{1}{T}}}.
\end{align*}
\end{customlemma}
See proof in Appendix \ref{proof:t_scale}.
In our experiments, $T=20$.
We split KL-divergence into two summands:
\begin{equation*}
    \operatorname{D}_{\mathrm{KL}}(p_T(c|\vecX) \| p(c)) = \mathrm{KL}_1 + \mathrm{KL}_2
\end{equation*}
where $\mathrm{KL}_1, \mathrm{KL}_2$ correspond to the two terms in Lemma~\ref{lemma:t_scale}.
The second term in KL-divergence is proportional to predicted concentration $\kappa({\vecX})$ for sample $\vecX$, thus unbounded. 
To remedy this, we normalize KL-divergence using the validation set:
\begin{equation*}
    \mathrm{KL}_i^{\mathrm{norm}} = \frac{\mathrm{KL}_i - \mathrm{KL}_i^{\mathrm{mean}}}{\mathrm{KL}_i^{\mathrm{std}}},\quad i \in \{1, 2\},
\end{equation*}
where $\mathrm{KL}_i^{\mathrm{mean}}$ and $\mathrm{KL}_i^{\mathrm{std}}$ are mean and std values of $i$-th component of KL-divergence computed on validation set.
We also perform further alignment using MLP applied to two summands:
\begin{equation}\label{eq:holue_mlp}
     q_{\mathrm{HolUE}} = f_{\boldsymbol{\theta}}\left(\mathrm{KL}_1^{\mathrm{norm}}, \mathrm{KL}_2^{\mathrm{norm}}\right)\in [0, 1],
\end{equation}
where $f_{\boldsymbol{\theta}}$ is Multilayer perceptron (MLP).
Parameters of MLP, $\boldsymbol{\theta}$, are trained on the validation set, solving a binary classification task: error/not error for a given false positive rate.

Additionally, we investigate direct normalized sum to show the necessity of MLP post-processing:
\begin{equation}\label{eq:holue_sum}
    q_{\mathrm{HolUE}} = \mathrm{KL}_1^{\mathrm{norm}} + \mathrm{KL}_2^{\mathrm{norm}}.
\end{equation}

\subsubsection{Summary}
We propose an uncertainty estimation method $\mathrm{HolUE}$ based on the Bayesian modeling for the OSR task.
The galley-aware method $\mathrm{GalUE}$ is combined with probabilistic embedding model $\mathrm{SCF}$ to merge two sources of uncertainty: uncertainty related to the relative position of embeddings and embedding variance of a biometric sample.
By convolving them, we obtain the posterior class distribution $p(c|\vecX)$ using  \eqref{eq:z_marginalization} and compare it to the uniform prior distribution $p(c)$ via KL-divergence to obtain an uncertainty measure.
Then, our approach performs temperature scaling of the posterior distribution for numerical stability and normalization of the two summands in the KL-divergence formula with the help of a validation set to adjust the scale of uncertainties from different sources.



%% file: parts/experiments.tex
\section{Experiments}
\label{sec:experiments}
In this section, we report the results of our numerical experiments, including additional ablation and sensitivity studies.
While $\mathrm{SCF}$~\cite{scf}, $\mathrm{ScaleFace}$~\cite{Kail2022ScaleFaceUD} and $\mathrm{PFE}$~\cite{pfe} allow for quality-aware template pooling, our study focuses mainly on risk-controlled OSR, where we filter probe samples which have high uncertainty.
We aim to be close to the best filtering procedure that drops erroneous examples.

\subsection{Compared methods}
We compare the following uncertainty estimation approaches $\mathrm{PFE}$, $\mathrm{SCF}$,  $\mathrm{ScaleFace(SF)}$ (Section~\ref{sec:embedding_uncertainty}), $\mathrm{AccScr}$ (Section~\ref{sec:pairwise_distance}) with two our methods, $\mathrm{GalUE}$ (Section~\ref{sec:deterministic_embedding}) and $\mathrm{HolUE}$ (Section~\ref{sec:combined_ue}).
The methods $\mathrm{PFE}$, $\mathrm{SCF}$, $\mathrm{SF}$ estimate only sample quality with the help of probabilistic embeddings, and we use them as baselines.
Method $\mathrm{AccScr}$ is a simple ad-hoc uncertainty measure based on an acceptance score.
It yields high uncertainty when the deterministic embedding of the probe sample lies close to the decision boundary between acceptance and rejection.
Our method $\mathrm{GalUE}$ also works with deterministic embeddings. In addition to the capabilities of $\mathrm{AccScr}$, it can capture ambiguity related to the position of embeddings in the gallery.
Finally, our approach $\mathrm{HolUE}$ enhances $\mathrm{GalUE}$ with probabilistic embeddings of $\mathrm{SCF}$ method.


\subsection{Metrics}
\paragraph{OSR accuracy}
We compute common metrics for OSR prediction accuracy estimation: false positive identification rate ($\mathrm{FPIR}$), false negative identification rate ($\mathrm{FNIR}$) and $F_1$ score.
We aim to maximize $F_1$ and minimize $\mathrm{FPIR}$, $\mathrm{FNIR}$.
See Appendix~\ref{sec:metrics} for strict definitions and discussions.

\paragraph{Uncertainty estimation quality}\label{sec:ua_metric}
A natural metric for evaluating the effectiveness of obtained quality scores is the area under a rejection curve~\cite{fadeeva2023lm}.
We filter probe samples with quality scores lower than some threshold $t$ and compute recognition accuracy metrics, e.g., $F_1$ score on the remaining test part.
So, methods that yield higher gains in target metrics after filtering are better.
The prediction rejection (PR) curve is constructed by varying the percentage of the rejected templates and recording a recognition metric.
The area under the PR curve is a quality measure: the higher the curve, the better we identify erroneous probe templates~\cite{Kail2022ScaleFaceUD}.
We normalize it, obtaining the Prediction Rejection Ratio (PRR)~\cite{fadeeva2023lm}, used for comparison:
$$
    \mathrm{PRR} = \frac{\mathrm{AUCPR}_{\mathrm{unc}} - \mathrm{AUCPR}_{\mathrm{random}}}{\mathrm{AUCPR}_{\mathrm{oracle}} - \mathrm{AURCPR}_{\mathrm{random}}}
,$$
where $\mathrm{AUCPR}_{\mathrm{unc}}$ -- area under the PR curve for an uncertainty estimation method; $\mathrm{AUCPR}_{\mathrm{random}}, \mathrm{AUCPR}_{\mathrm{oracle}}$ -- areas under random and optimal PR curves correspondingly. 
$\mathrm{Random}$ filters probe samples in random order and $\mathrm{Oracle}$ filters erroneous samples first.

\subsection{Datasets}

Common IJB-B~\cite{ijbb} and IJB-C~\cite{ijbc} datasets, universally acknowledged for such purposes in the face recognition literature, lay a foundation for our comparison.
To broaden the list of problems we explore audio domain and test our method on OSR dataset VoxBlink~\cite{voxblink}.
We test out methods on VB-Eval-Large-5 with $5$ enrollment utterances per speaker in gallery, and use VB-Eval-M-5 as the validation set. 
We also add a new OSR protocol, we call \textit{Whale}, on whale identity prediction.
The dataset is based on the Happawhale Kaggle competition~\cite{happy-whale-and-dolphin} and has $15587$ identities and $51033$ images.
To conduct a comparison, we require retrained backbone models and a recognition protocol.
ArcFace~\cite{deng2019arcface} and $\mathrm{SCF}$~\cite{scf} serve as backbone models with their checkpoints available in our \href{https://anonymous.4open.science/r/face_ue-64D4/README.md}{repo}.
We also use a subset of MS1MV2~\cite{deng2019arcface} dataset for validation.
The detailed OSR protocols are presented in Appendix~\ref{sec:whale_osr},~\ref{sec:validation_set}.
\subsection{Main results}

\paragraph{Aggregated metrics} Table~\ref{tab:filter} presents PRR for the $F_1$ metric.
Each method starts at the same $\mathrm{FPIR}$, and we drop $50\%$ test samples to compute PRR.
Results for three datasets and different $\mathrm{FPIR}$ values are in Table~\ref{tab:filter}.

\begin{figure}[!b]
\newcommand\figuresize{0.235}
    \begin{subfigure}[b]{\figuresize\textwidth}
\includegraphics[width=\textwidth]{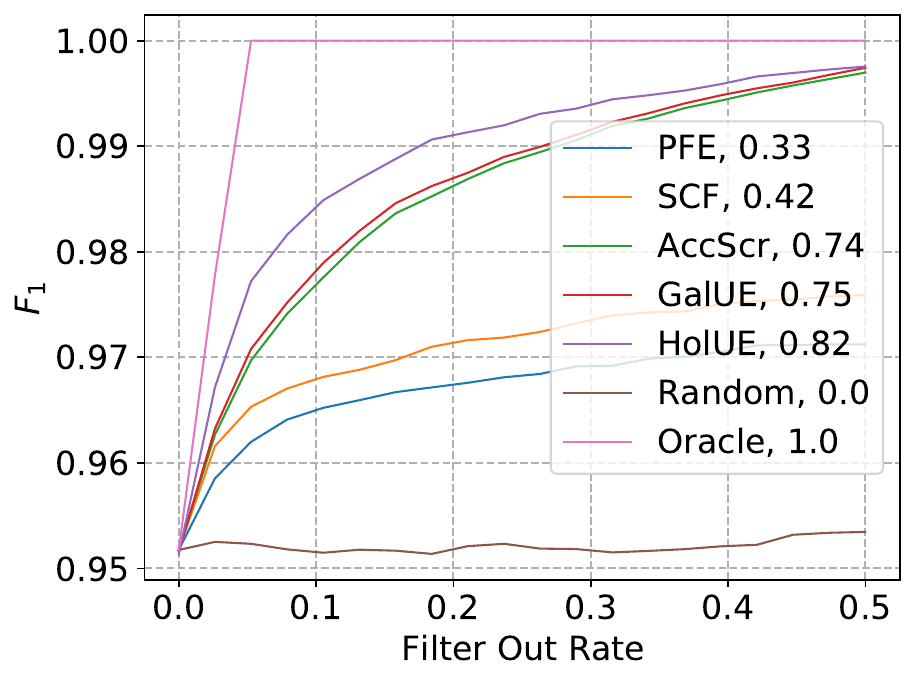}
    \centering
    \caption{$F_1 \uparrow$}
\label{fig:risk_control_filter}
    \end{subfigure}
    \hfill
    \begin{subfigure}[b]{\figuresize\textwidth}
\includegraphics[width=\textwidth]{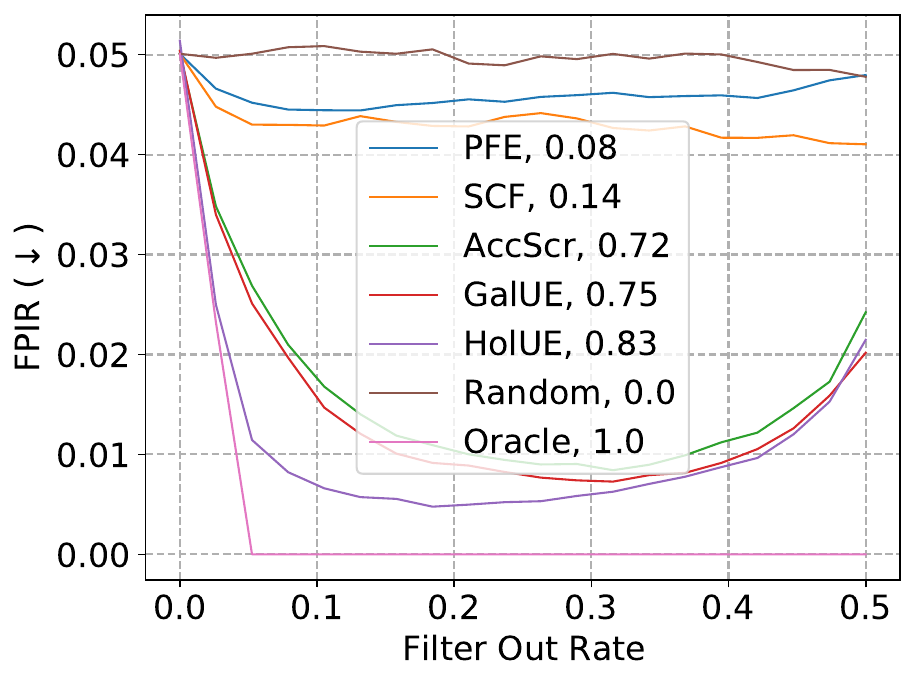}
    \centering
    \caption{$\mathrm{FPIR} \downarrow$}
    \label{fig:fpir_filter}
    \end{subfigure}
    \hfill
    \begin{subfigure}[b]{\figuresize\textwidth}
\includegraphics[width=\textwidth]{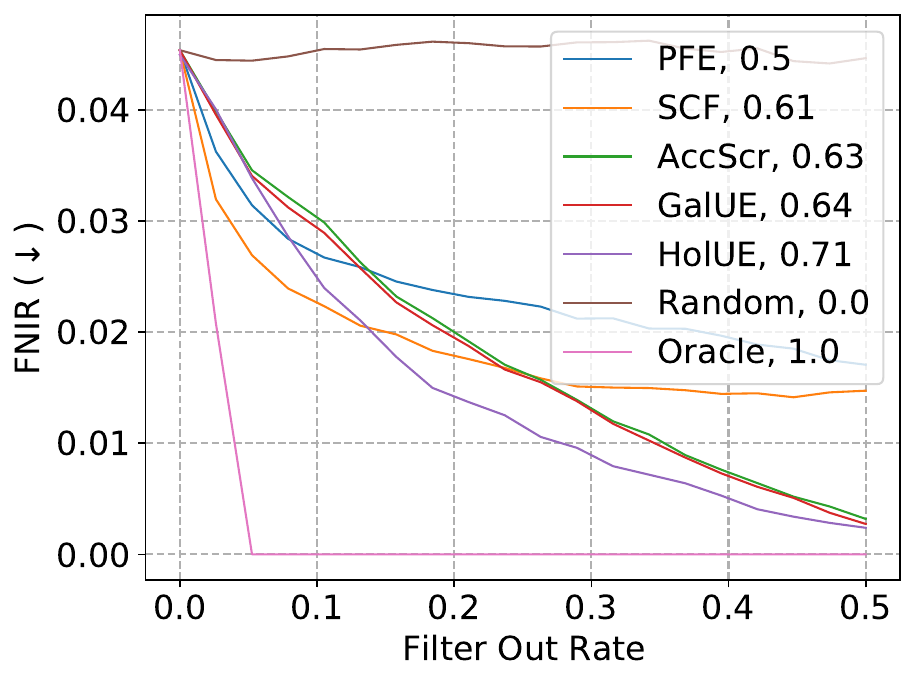}
    \centering
    \caption{$\mathrm{FNIR} \downarrow$}
    \label{fig:fnir_filter}
    \end{subfigure}
    \hfill
    \begin{subfigure}[b]{\figuresize\textwidth}
\includegraphics[width=\textwidth]{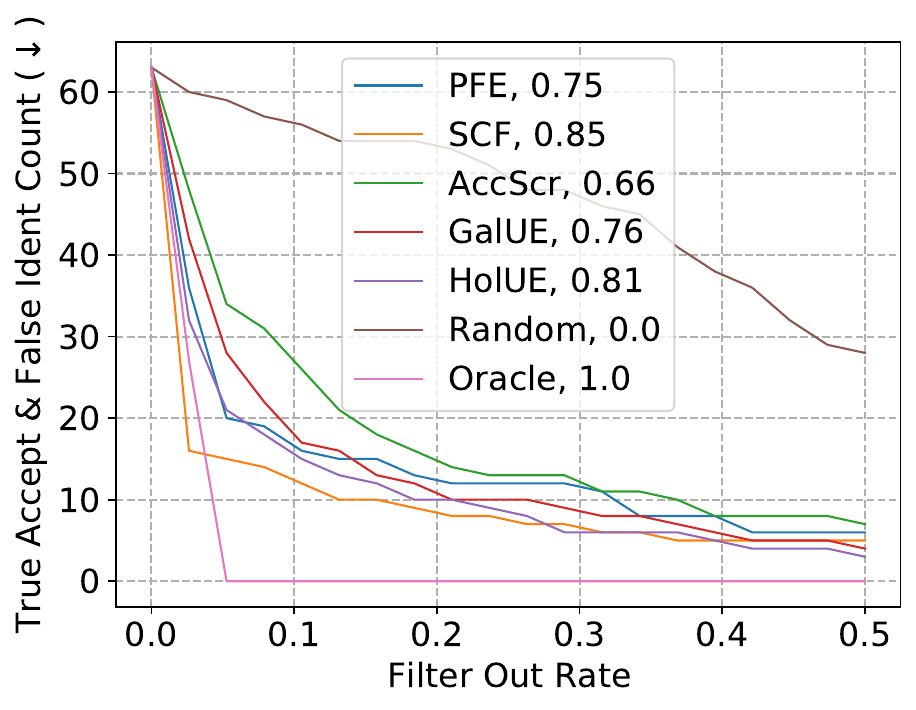}
    \centering
    \caption{$\mathrm{Identity\,\, count} \downarrow$}
    \label{fig:ident_filter}
    \end{subfigure}
    \hfill
    \caption{Risk-Controlled Open-set Face Recognition on dataset IJB-C starting with $\mathrm{FPIR} = 0.05$.
    Prediction Rejection Ratio is reported. Better to view in zoom.}
    \label{fig:main_rejection_curves}
\end{figure}

\paragraph{Rejection curves} 
We draw rejection curves based on different uncertainty estimates.
The backbone model is the same, so we use the same average embedding vectors for templates, and all curves start from the same point of $\mathrm{FPIR}$ for a zero filtering rate.
The rejection curves for the IJB-C dataset for four considered metrics are presented in Figures~\ref{fig:main_rejection_curves}.
The reference PR curves for $\mathrm{Random}$ and $\mathrm{Oracle}$ rejection give additional context.

\begin{table}[t]
\caption{Prediction Rejection Ratios (PRR, $\uparrow$) for $F_1$ filtering curve for the four considered datasets IJB-C, IJB-B, Whale, VB-Eval-L.
}
\label{tab:filter}
\footnotesize
\centering
\setlength\tabcolsep{2pt}
\begin{tabular}{lcccccccccccc}
\toprule
Method & \multicolumn{12}{c}{$\mathrm{FPIR}$} \\
 & $0.05$ & $0.1$ & $0.2$ & $0.05$ & $0.1$ & $0.2$ & $0.05$ & $0.1$ & $0.2$ & $0.01$ & $0.05$ & $0.1$ \\
\midrule
 & \multicolumn{3}{c}{IJB-C} & \multicolumn{3}{c}{IJB-B} & \multicolumn{3}{c}{Whale} & \multicolumn{3}{c}{VB-Eval-L-5} \\
\midrule
PFE &  0.38 &  0.26 &  0.14 &  0.24 &  0.2 &  0.19 &  0.0 &  -0.01 &  -0.01 & 0.56 & 0.33 & 0.2 \\
SCF &  0.42 &  0.32 &  0.22 &  0.29 &  0.25 &  0.22 &  0.16 &  0.01 &  -0.1 & 0.55 & 0.27 & 0.13 \\
SF & 0.38 & 0.27 & 0.14 & 0.3 &0.21 & 0.16 & 0.19 & 0.11 & 0.03 & 0.44 & 0.52 & 0.44 \\
AccScr &  0.74 &  0.73 &  0.66 &  0.65 & \underline{0.69} &  0.63 & \underline{0.77} &  0.75 &  0.65 & \underline{0.65} & 0.77 & 0.73 \\
GalUE & \underline{0.75} & \underline{0.76} & \underline{0.7} & \underline{0.67} & \textbf{0.7} & \underline{0.66} &  0.74 & \underline{0.76} & \underline{0.7} & 0.64 & \textbf{0.89} & \underline{0.88} \\
HolUE & \textbf{0.82} & \textbf{0.79} & \textbf{0.87} & \textbf{0.71} &  0.58 & \textbf{0.82} & \textbf{0.82} & \textbf{0.87} & \textbf{0.9} & \textbf{0.85} & \underline{0.86} & \textbf{0.92} \\
\bottomrule
\end{tabular}
\end{table}

\paragraph{Analysis}
Our methods $\mathrm{GalUE}$ and $\mathrm{HolUE}$ outperform sample quality-based uncertainty estimators, $\mathrm{SCF}$, $\mathrm{PFE}$ and simple ad-hoc rejection/acceptance boundary-aware $\mathrm{AccScr}$ in all considered datasets.
Additional metrics provided in Figure~\ref{fig:main_rejection_curves} sheds light on the methodological problems of the baselines.
$\mathrm{SCF}$ and $\mathrm{PFE}$ underperform, as they poorly detect non-mated probe sample that are close to some subjects in a gallery, which would occur if the test size is vast.
The quality of such probe subjects may be high, and thus, the FPIR will reduce slowly as we filter test samples (see Figire~\ref{fig:fpir_filter}). 
Contrarily, $\mathrm{GalUE}$ and $\mathrm{AccScr}$ detect false accepted samples because the acceptance score of such samples is close to the accept/reject decision boundary due to their imperfect similarity to some subjects in the gallery.
On the other hand, $\mathrm{SCF}$ accurately detects the false reject error type because it mainly occurs when an sample is of poor quality (see Figire~\ref{fig:fnir_filter}).
The $\mathrm{GalUE}$ and $\mathrm{AccScr}$ struggle with such cases because they classify such samples as out-of-gallery with high confidence.
Among these two methods, $\mathrm{GalUE}$ provides a higher PR curve than the $\mathrm{AccScr}$ at detecting identification errors (see Figure~\ref{fig:ident_filter}).
This behaviour is expected because unlike $\mathrm{AccScr}$, $\mathrm{GalUE}$ considers information about the position of test sample embeddings' relative to gallery classes.
Method $\mathrm{HolUE}$ unites strong sides of $\mathrm{SCF}$ and $\mathrm{GalUE}$: the $\mathrm{SCF}$ part of the score identifies false rejects, while the $\mathrm{GalUE}$ part detects false accepts.
\subsection{Ablation studies}
\paragraph{Post-processing}
In Table~\ref{tab:filter_ablation} we compare post-processing strategies described in Section~\ref{sec:combined_ue}. 
HolUE-sum designated normalized sum defined in~\eqref{eq:holue_sum}, and HolUE-MLP is defined in~\eqref{eq:holue_mlp}.
The "val" and "test" markers point to a dataset on which the mean and standard deviation values for normalization were computed. 
The MLP was trained on validation set in both cases.
Fair normalization using the validation dataset is sufficient to get high PRR values.
We see, that MLP head provides quality boost and allows for better performance when only a validation set is available.

\begin{figure}
    \newcommand\figuresize{0.95}
     \centering
\subfloat{%
\includegraphics[clip,width=\figuresize\columnwidth]{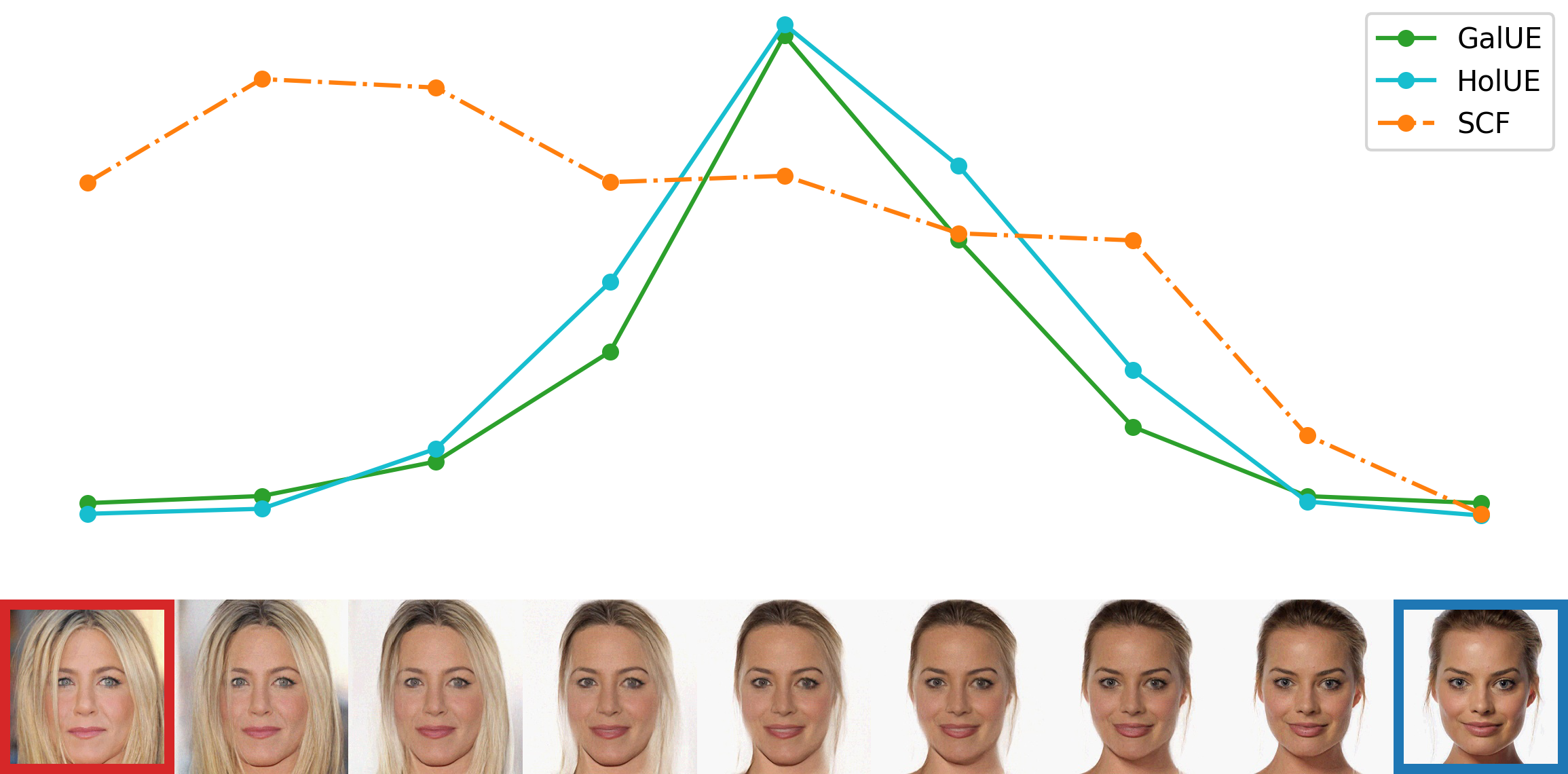}
}\caption{
Uncertainty score comparison (higher values correspond to higher uncertainty) of $\mathrm{SCF}$ with our methods $\mathrm{GalUE}$ and $\mathrm{HolUE}$  for a toy face interpolation example.
    Images in between serve as probe images.
}
    \label{fig:unc_iterpolate_main}
\end{figure}

\begin{table}[!t]
\caption{Prediction Rejection Ratio (PRR) of $F_1$ filtering curve on IJB-C dataset.
}
\label{tab:filter_ablation}
\small
\centering
\setlength\tabcolsep{6pt}
\begin{tabular}{lccc}
\toprule
Method & \multicolumn{3}{c}{$\mathrm{FPIR}$} \\
 & $0.05$ & $0.1$ & $0.2$ \\
\midrule
HolUE-sum test &  0.74 &  0.74 &  0.74 \\
HolUE-sum val &  0.52 &  0.26 &  0.47 \\
HolUE-MLP test & \textbf{0.82} & \textbf{0.84} & \textbf{0.87} \\
HolUE-MLP val & \textbf{0.82} & \underline{0.79} & \textbf{0.87} \\
\bottomrule
\end{tabular}
\end{table}
\paragraph{Qualitative analysis}
We also provide illustrative examples to grasp the capabilities of our gallery-aware method and prove that it produces reasonable uncertainty scores.
Figure \ref{fig:unc_iterpolate_main} presents results for a simple OSR problem where two images of different people are used as gallery images, and interpolated images serve as probe examples.
Red and blue frames designate gallery images.
We use interpolated images from the \textit{Glow} article~\cite{NEURIPS2018_d139db6a}.
For most blended and thus ambiguous probe images in the middle, our methods $\mathrm{GalUE}$ and $\mathrm{HolUE}$ yield the highest uncertainty value.
The $\mathrm{SCF}$-based data uncertainty notices only the quality of probe images (hair covers part of the face of Jennifer Aniston) ignoring the embeddings relative position and thus cannot detect decision ambiguity of images in the middle. 
Uncertainty scores are normalized independently to be of the same scale.

%% file: parts/conclusion.tex
\section{Conclusion}
\label{sec:conclusion}
We presented a method for uncertainty estimation in open-set recognition that distinguishes itself from others by providing a way to cover multiple sources of uncertainty.
Our gallery-aware method for the open-set recognition problem provides a fully probabilistic treatment of the problem at hand and gives a natural uncertainty estimate.
Combining the gallery-aware uncertainty estimate with a measure of biometric sample quality yields additional fundamental improvement, as the latter accounts for false rejection errors.
The core mechanism for aggregation is the approximate computation of posterior class distribution.
When applied to widely used datasets in the face and audio recognition literature, our method shows a clear improvement in risk-controlled metrics over simple sample quality-based filtering, providing evidence of its superiority.
We also provide an open-set recognition protocol for the Whale dataset and show that our method is superior in this challenging task as well.

%% file: parts/appendix.tex
\appendix

\section{Appendix}

The appendix contains the following sections that complement the main body of the paper:
\begin{itemize}
    \item definition of the OSR quality metrics
    \item proof of the statements given in the paper.
    \item exact computation of posterior probabilities;
    \item proof that our gallery-aware uncertainty $\mathrm{GalUE}$ is equivalent to a score function-based baseline in terms of OSR predictions;
    \item description of the Whale dataset, validation set, and corresponding testing protocols;
\end{itemize}    

\subsection{Quality metrics for uncertainty evaluation in OSR}\label{sec:metrics}
\label{sec:evaluation}
\paragraph{OSR quality}
To define OSR errors, three sets of  template sets are defined:
\begin{itemize}
    \item $\mathcal{G}$ --- a enrolment database,
    \item $\mathcal{P}_{\mathcal{G}}$ --- a set of probes which have a corresponding mated class in the enrolment database,
    \item $\mathcal{P}_{\mathcal{N}}$ --- a set of probes which have no corresponding class in the gallery.
\end{itemize}
For each probe in $\mathcal{P}_{\mathcal{G}}$ and $ \mathcal{P}_{\mathcal{N}}$, the OSR predictions described in the previous section are produced.
The OSR system should reject probes from $\mathcal{P}_{\mathcal{N}}$ set, accept and correctly identify probes from $\mathcal{P}_{\mathcal{G}}$ set.
So, we can have three kinds of OSR errors: False Acceptance, False Rejection, and Misidentification.

We use false positive identification rate (FPIR) and false negative identification rate (FNIR) to measure identification performance:
\begin{align*}
    &\mathrm{FPIR} = \frac{\mathrm{FP}}{|\mathcal{P}_{\mathcal{N}}|}\\
    &\mathrm{FNIR} = \frac{\mathrm{FN}}{\mathrm{FN}+\mathrm{TP}}
\end{align*}
where a probe $p$ from $\mathcal{P}_{\mathcal{G}}$ is True Positive ($\mathrm{TP}$), if $p \text{ is accepted and}\, \,\hat{\id}(p) = \id(p)$, and False Negative ($\mathrm{FN}$) otherwise.
We count probe $p$ as False Positive ($\mathrm{FP}$), when $p\in \mathcal{P}_{\mathcal{N}}, \text{and } p \text{ is accepted}$.
In the formulae, we refer to abbreviations $\mathrm{TP}$, $\mathrm{FN}$, and $\mathrm{FP}$ as the number of objects in the corresponding sets.
These metrics are related to metrics with old-style notation false accept rate ($\mathrm{FAR}$) and detection\&identification rate ($\mathrm{DIR}$): $\mathrm{FAR} = \mathrm{FPIR}$ and $\mathrm{DIR} = 1-\mathrm{FNIR}$~\cite{face_handbook}.
To get a unified recognition performance measure, we also compute the $F_1$ score:
\begin{equation*}
    \mathrm{F}_1 = 2 \frac{\mathrm{precision} \cdot \mathrm{recall}}{\mathrm{precision} + \mathrm{recall}},
\end{equation*}
\begin{equation*}
    \mathrm{precision} = \frac{\mathrm{TP}}{\mathrm{TP} + \mathrm{FP}},\quad \mathrm{recall} = \frac{\mathrm{TP}}{\mathrm{TP} + \mathrm{FN}}, 
\end{equation*}

We maximize $\mathrm{F}_1$, and minimize $\mathrm{FPIR}$ and $\mathrm{FNIR}$.
We also can fix $\mathrm{FPIR}$ by varying a threshold and report $\mathrm{FNIR}$ or another metric for a fixed $\mathrm{FPIR}$.

\subsection{Proofs}\label{sec:proofs}
\begin{customlemma}{1}\label{proof:p_z}
With $p(c)$ and $p(\vecZ|c)$ defined in Section \ref{sec:deterministic_embedding}
    \begin{equation*}
        p(\vecZ) = \frac{1-\beta}{K}\sum_{i=1}^{K}p(\vecZ|c=i) + \frac{\beta}{S_{d-1}}
    \end{equation*}
\end{customlemma}
\begin{proof}
    \begin{align*}
    &p(\vecZ)=\int_{0}^{K+1} p(\vecZ|c)p(c)dc =\frac{1-\beta}{K}\sum_{i=1}^{K}p(\vecZ|c=i) +\\
    &+ \int_{K}^{K+1}p(\vecZ|c)p(c)dc\\
    &\int_{K}^{K+1}p(\vecZ|c)p(c)dc = \beta\int_{\mathbb{S}_{d - 1}}\delta \left(\vecZ - \vecM_c^{\circ}\right)\frac{d\vecM_c^{\circ}}{S_{d-1}} =
    \frac{\beta}{S_{d-1}}\\
    &p(\vecZ) = \frac{1-\beta}{K}\sum_{i=1}^{K}p(\vecZ|c=i) + \frac{\beta}{S_{d-1}},\quad S_{d-1} = \frac{2\,\pi^\frac{d}{2}}{\Gamma\left(\frac{d}{2}\right)}
\end{align*}
where $S_{d-1}$ is area of d-dimensional unit sphere and $\Gamma$ is Gamma function.
\end{proof}

\begin{customlemma}{2}\label{proof:t_scale}
    KL-divergence after scaling with temperature T has the following form:
    \begin{align*}
        &\operatorname{D}_{\text{KL}}(p_T(c|\vecX) \| p(c))=\sum_{c=1}^K \mathbb{P}_T(c|\vecX) \log \frac{\mathbb{P}_T(c|\vecX)}{\mathbb{P}(c)}+ \\
+&\frac{\beta^{\frac{1}{T}}}{S_{d-1}^{\frac{1}{T}}} \frac{1}{p(\vecM_{\vecX})} \log\left[\left(\frac{\beta}{S_{d-1}}\right)^{\frac{1}{T}-1}\frac{p\left(\vecM_{\vecX}|\vecX\right)}{p\left(\vecM_{\vecX}\right)}\right]\\
&\mathbb{P}_T(c|\vecX) = \frac{\left(\frac{1-\beta}{K}\right)^{\frac{1}{T}}p(\vecM_{\vecX}|c)^{\frac{1}{T}}}{\left(\frac{1-\beta}{K}\right)^{\frac{1}{T}}\sum\limits_{c=1}^{K}p(\vecM_{\vecX}|c)^{\frac{1}{T}} + \left(\frac{\beta}{S_{d-1}}\right)^{\frac{1}{T}}}
    \end{align*}
\end{customlemma}

\begin{proof}
We want to find a form for temperature scaling that is similar to scaling of discrete distribution and, with $T=1$, gives original formulas.
For gallery classes, we use standard temperature scaling:
\begin{align}
    p(c)p(\vecZ|c)\rightarrow p(c)^{\frac{1}{T}}p(\vecZ|c)^{\frac{1}{T}}, c\in\{1,\dots,K\}
\end{align}
For out-of-gallery classes with $c\in(K,K+1]$:
\begin{equation*}
        p(\vecZ|c)=\delta \left(\vecZ - \vecM_c^{\circ}\right) \rightarrow \frac{S_{d-1}}{S_{d-1}^{\frac{1}{T}}}\delta \left(\vecZ - \vecM_c^{\circ}\right)=p_T(\vecZ|c) 
    \end{equation*}
With this
\begin{equation*}
    p_T(\vecZ) = \left(\frac{1-\beta}{K}\right)^{\frac{1}{T}}\sum\limits_{c=1}^{K}p(\vecZ|c)^{\frac{1}{T}} + \left(\frac{\beta}{S_{d-1}}\right)^{\frac{1}{T}}
\end{equation*}
With $T=1$, we get the original unscaled distribution.
Let us find the scaled posterior distribution:
\begin{align*}
    &p_{T}(c|\vecX) =  \int_{\mathbb{S}_{d-1}}\frac{S_{d-1}}{S_{d-1}^{\frac{1}{T}}}\frac{\delta(\vecZ-\vecM_c^{\circ})\beta^{\frac{1}{T}}}{p(\vecZ)}p(\vecZ|\vecX)d\vecZ=\\
    &=\frac{S_{d-1}}{S_{d-1}^{\frac{1}{T}}}\beta^{\frac{1}{T}}\frac{p(\vecM_c^{\circ}|\vecX)}{p(\vecM_c^{\circ})}, c\in(K,K+1]\\
    &\mathbb{P}_T(c|\vecX) = \int_{\mathbb{S}_{d - 1}}\left( \frac{1 - \beta}{K}\right)^{\frac{1}{T}}\frac{p_T(\vecZ|c)}{p_T(\vecZ)}p(\vecZ|\vecX) d\vecZ.\\
     &\mathbb{P}_T(c|\vecX)\approx \left( \frac{1 - \beta}{K}\right)^{\frac{1}{T}}\frac{p_T(\vecM_{\vecX}|c)}{p_T(\vecM_{\vecX})}, c\in\{1,\dots,K\}
\end{align*}
Now we can compute KL-divergence:
\begin{align*}
    &\operatorname{D}_{\text{KL}}(p_T(c|\vecX) \| p(c)) = \sum_{c=1}^K \mathbb{P}_T(c|\vecX) \log \frac{\mathbb{P}_T(c|\vecX)}{\mathbb{P}(c)}+\\
    &+\int_K^{K+1} p_T(c|\vecX) \log \frac{p_T(c|\vecX)}{p(c)} dc\\
    &\int_K^{K+1} p_T(c|\vecX) \log \frac{p_T(c|\vecX)}{p(c)} dc  =\\&
    =\int_{\mathbb{S}_{d - 1}}\frac{S_{d-1}\beta^{\frac{1}{T}}}{S_{d-1}^{\frac{1}{T}}}\log\left[\frac{\beta^{\frac{1}{T}-1}}{S_{d-1}^{\frac{1}{T}-1}}\frac{p\left(\vecM_c^\circ| \vecX\right)}{p\left(\vecM_c^\circ\right)}\right]\frac{p\left(\vecM_c^\circ| \vecX\right)}{p\left(\vecM_c^\circ\right)}\frac{d \vecM_c^\circ}{S_{d-1}}\approx\\
    &\approx \frac{\beta^{\frac{1}{T}}}{S_{d-1}^{\frac{1}{T}}} \frac{1}{p(\vecM_{\vecX})} \log\left[\left(\frac{\beta}{S_{d-1}}\right)^{\frac{1}{T}-1}\frac{p\left(\vecM_{\vecX}|\vecX\right)}{p\left(\vecM_{\vecX}\right)}\right]
\end{align*}
\end{proof}

\subsection{Probability computation}\label{sec:computation}
To avoid possible numerical errors, we first compute the logarithm of probability and then exponentiate it.
Below, we provide formulas for the $\mathrm{GalUE}$ method. 

From the Bayes formula:
\begin{equation*}
    \log \mathbb{P}(c|\vecZ) = \log p(\vecZ|c) + \log \mathbb{P}(c) - \log p(\vecZ)
\end{equation*}

The first two terms were defined in the main body of the paper.
For the last term, we will use the technical lemma below.

\begin{customlemma}{3}
    For $p(\vecZ)$ we have the following form:
    \begin{align*}
    \log p(\vecZ) = \log\left[\frac{1-\beta}{K}\sum_{c=1}^K \exp\left(\kappa\vecM^T_{c}\vecZ\right) + {}_0F_1\left(;n;\frac{\kappa^2}{4}\right)\beta\right],
\end{align*}
where ${}_0F_1$ is the generalized hypergeometric function.
\end{customlemma}
\begin{proof}
The last term is (see Lemma \ref{lem:p_z}):
\begin{align*}
&p(\vecZ)=\frac{1-\beta}{K}\sum_{i=1}^{K}p(\vecZ|c=i) + \frac{\beta}{S_{d-1}}=\\
&=\sum_{c=1}^K\mathcal{C}_d(\kappa)\exp\left(\kappa\vecM^T_{c}\vecZ\right)p(c) + \frac{\beta}{S_{d-1}}=\\
&=\mathcal{C}_d(\kappa)\left[\sum_{c=1}^K\exp\left(\kappa\vecM^T_{c}\vecZ\right)p(c) + \frac{1}{S_{d-1}\mathcal{C}_d(\kappa)}\beta\right]
\end{align*}
let $\alpha(\kappa) = \frac{1}{S_{d-1}\mathcal{C}_d(\kappa)}$, then
\begin{equation*}
    \log p(\vecZ) = \log \mathcal{C}_d(\kappa) + \log\left[\sum_{c=1}^K\exp\left(\kappa\vecM^T_{c}\vecZ\right)p(c) + \alpha(\kappa)\beta\right]
\end{equation*}
\begin{equation*}
    \alpha(\kappa) = \frac{\Gamma\left(\frac{d}{2}\right)(2\pi)^{d/2}\mathcal{I}_{d/2-1}(\kappa)}{2\pi^{\frac{d}{2}}(\kappa)^{d/2-1}}
\end{equation*}
Let $n = \frac{d}{2}$,
\begin{align*}
    &\alpha(\kappa) = \frac{\Gamma\left(n\right)(2\pi)^{n}\mathcal{I}_{n-1}(\kappa)}{2\pi^{n}(\kappa)^{n-1}} = \frac{(n-1)!2^{n-1}}{\kappa^{n-1}}\mathcal{I}_{n-1}(\kappa)\\
    &\mathcal{I}_{n-1}(\kappa) = \sum_{m=0}^{\infty}\frac{1}{m!\Gamma(m+n)}\left(\frac{\kappa}{2}\right)^{2m+n-1}\\
    &\alpha(\kappa) = \sum_{m=0}^{\infty}\frac{(n-1)!}{m!(m+n-1)!}\left(\frac{\kappa}{2}\right)^{2m} = \sum_{m=0}^{\infty}\frac{1}{m!(n)_m}\left(\frac{\kappa^2}{4}\right)^{m}
\end{align*}
Here, the notation of rising factorial is used (Pochhammer symbol)
\begin{align*}
(a)_0 &= 1, \\
(a)_n &= a(a+1)(a+2) \cdots (a+n-1), && n \geq 1
\end{align*}
We can see that $\alpha(\kappa) = {}_0F_1\left(;n;\frac{\kappa^2}{4}\right)$.
\begin{align*}
    \log \mathcal{C}_d(\kappa) = \left[\frac{d}{2}-1\right]\log \kappa - \frac{d}{2}\log 2\pi - \log \mathcal{I}_{d/2-1}(\kappa)
\end{align*}
Finally
\begin{align*}
    &\log p(\vecZ) = (n-1)\log\kappa - n\log 2\pi - \log\mathcal{I}_{n-1}(\kappa)+\\
&+\log\left[\sum_{c=1}^K\frac{\exp\left(\kappa\vecM^T_{c}\vecZ\right)(1-\beta)}{K} + {}_0F_1\left(;n;\frac{\kappa^2}{4}\right)\beta\right]
\end{align*}
\end{proof}

For gallery class with label $c\in\{1,\dots,K\}$, we have:
\begin{align*}
    \log \mathbb{P}(c|\vecZ) = \log \mathcal{C}_d(\kappa) + \kappa\vecM^T_{c}\vecZ + \log \left(\frac{1-\beta}{K}\right) - \log p(\vecZ),
\end{align*}
where $p(\vecZ)$ is presented above.

\subsection{Recognition decision equivalence}
\label{sec:score_function_connection}
This section proves that the score function-based baseline (Section~\ref{sec:pairwise_distance}) and gallery-aware method $\mathrm{GalUE}$ (Section~\ref{sec:deterministic_embedding}) yield the same OSR predictions when appropriately tuned.
Here, we denote these algorithms $A1$ and $A2$ respectively.
Below, we prove that there exists $(\tau, \kappa, \beta)$ such that on all test templates $p$, both methods produce the same answer. 
First, let us prove that if  $p$ is accepted, both methods will predict the same gallery class. 
\begin{align*}
    &\hat{\id}_{A2}(p) = \argmax_{c \in \{1, \ldots, K\}} \mathbb{P}(c|\vecZ) = \\
    &= \argmax_{c \in \{1, \ldots, K\}} \frac{\mathcal{C}_d(\kappa) \exp\left(\kappa\vecM^T_{c} \vecZ\right)\mathbb{P}(c)}{p(\vecZ)} =\\
    &= \argmax_{c \in \{1, \ldots, K\}} \vecM^T_{c} \vecZ = \hat{\id}_{A1}(p).
\end{align*}

Let's prove that $A2$ accepts $p \iff A1$ accepts $p$, with 
\begin{equation*}
    \tau = \frac{1}{\kappa} \log\left(\frac{\beta}{1 - \beta} \frac{K}{S_{d - 1} \mathcal{C}_d(\kappa)} \right).
\end{equation*}

\begin{align*}
    &p \text{ is accepted by A2} \iff \\
    &\iff \frac{\beta}{S_{d-1}} \leq \max_{c \in \{1, \ldots, K\}} p(\vecZ|c) \mathbb{P}(c) \\
    &\iff \max_{c \in \{1, \ldots, K\}} \frac{1 - \beta}{K} \mathcal{C}_d \exp \left(\kappa \vecM^T_{c} \vecZ \right) \geq \frac{\beta}{S_{d - 1}} \\
    &\iff \log\left(\frac{1 - \beta}{K} \mathcal{C}_d\right)  + \max_{c \in \{1, \ldots, K\}} \kappa \vecM^T_{c} \vecZ \geq \log \left(\frac{\beta}{S_{d - 1}}\right) \\
    &\iff \max_{c \in \{1, \ldots, K\}} \vecM^T_{c} \vecZ \geq \underbrace{\frac{1}{\kappa} \log\left(\frac{\beta}{1 -\beta} \frac{K}{S_{d - 1} \mathcal{C}_d(\kappa)}\right)}_{\tau}\\
    &\iff p\text{ is accepted by A1.}
\end{align*}
Although both methods produce the same recognition answers, our method is equipped with natural uncertainty estimation capability.

\subsection{Whale identification}
\label{sec:whale_osr}
We construct an Open Set Recognition (OSR) protocol for the Happywhale dataset \cite{happy-whale-and-dolphin} and use it to test our uncertainty estimation models on challenging non-human data.
Happywhale datasets train part has $51033$ images of $15587$ different animal identities, covering $30$ species.
We train the ArcFace-like model, following the first-place Kaggle solution \cite{patton2023deep}, to compute image embeddings.
On top of the embedding model, we train the $\mathrm{SCF}$~\cite{scf} and $\mathrm{PFE}$~\cite{pfe} models to estimate image quality.
Since the test dataset does not have identity labels, we built the OSR protocol using Happywhale train data.
We take $5000$ of $15587$ identities to create a validation set on which we train the calibration model.
The remaining $10587$ identities are used to test our uncertainty estimation methods.

We use the same splitting strategy for both validation and test sets to define gallery and test templates: IDs with one image are treated as out-of-gallery samples and others as in-gallery samples.
As a result, we have $2985$ out-of-gallery IDs and $2015$ in-gallery IDs for the validation set, $6273$ out-of-gallery IDs, and $4314$ in-gallery IDs for the test set.
We then randomly split images of each in-gallery identity into two templates: a gallery template and a test in-gallery template.

\subsection{MS1MV2 validation set}
\label{sec:validation_set}
For recognition in IJB-B and IJB-C datasets, we obtain validation dataset through the construction of OSR protocol using the MS1MV2 dataset with ArcFace loss function and ResNet model. The detailed protocol follows the one presented in \cite{deng2019arcface}. 
To construct the OSR protocol, we randomly sample $3531$ subjects (as in the IJB-C dataset) with more than $30$ images of available $87000$ subjects in MS1MV2.
Out of the $3531$ sampled subjects, we consider the first $1772$ known. We split their images into five random templates to construct four probe and one gallery template, and other subject images are used to get our-of-gallery probe templates. 